\documentclass{Interspeech}

\interspeechcameraready

\title{Representing Speech Through Autoregressive Prediction of Cochlear Tokens}

\author[affiliation={1}]{Greta}{Tuckute}
\author[affiliation={2}]{Klemen}{Kotar}
\author[affiliation={1,4}]{Evelina}{Fedorenko}
\author[affiliation={2,3}]{Daniel L. K.}{Yamins}

\affiliation{Department of Brain and Cognitive Sciences \& McGovern Institute for Brain Research}{MIT}{USA}
\affiliation{Department of Computer Science \& Wu Tsai Neurosciences Institute}{Stanford University}{USA}
\affiliation{Department of Psychology}{Stanford University}{USA}
\affiliation{Program in Speech and Hearing Bioscience and Technology}{Harvard University}{USA}
\email{gretatu@mit.edu, klemenk@stanford.edu, evelina9@mit.edu, yamins@stanford.edu}

\usepackage{comment}

\usepackage[T1]{fontenc}
\usepackage[utf8]{inputenc}
\usepackage[english]{babel}
\usepackage[T1]{fontenc}
\usepackage[utf8]{inputenc}
\usepackage[english]{babel}

\begin{document}

\maketitle

\begin{abstract}
We introduce \textbf{AuriStream}, a biologically inspired model for encoding speech via a two-stage framework inspired by the human auditory processing hierarchy. The first stage transforms raw audio into a time-frequency representation based on the human cochlea, from which we extract discrete \textbf{cochlear tokens}. The second stage applies an autoregressive sequence model over the cochlear tokens.
AuriStream learns meaningful phoneme and word representations, and state-of-the-art lexical semantics. AuriStream shows competitive performance on diverse downstream SUPERB speech tasks. Complementing AuriStream's strong representational capabilities, it generates continuations of audio which can be visualized in a spectrogram space and decoded back into audio, providing insights into the model's predictions.
In summary, we present a two-stage framework for speech representation learning to advance the development of more human-like models that efficiently handle a range of speech-based tasks\footnote{Website and model weights:

\mbox{\url{https://tukoresearch.github.io/auristream-speech/}}}

\end{abstract}

\section{Introduction}
\label{sec:intro}
Humans possess a remarkable ability to perform a wide range of tasks on speech inputs, from recognizing words in noise to separating speakers' voices and interpreting emotional tone. These processes are carried out by the human ear and networks of biological neurons. However, developing artificial neural networks that mirror the human ability to flexibly and efficiently understand and interact with the world through speech remains a significant challenge\cite{yang2021superb,wu2024codec,mehrish2023reviewdeeplearningtechniques}. \textbf{To bridge this gap, we introduce AuriStream, a biologically-inspired model that learns versatile speech representations through a simple and scalable autoregressive prediction objective on a time-frequency representation inspired by the human cochlea}.

\subsection{Related Work: Speech Representation Learning}
Speech representation models, also known as audio encoders, broadly transform audio signals into discrete tokens or continuous embeddings for various downstream audio tasks \cite{mehrish2023reviewdeeplearningtechniques}. \textbf{One popular approach is neural audio codecs}, which learn compressed representations by retaining the essential information for audio reconstruction, enabling them to recover the original signal from the learned codes \cite{Zeghidour2021SoundStreamAE,Defossez2022HighFN,yang2023hifi,kumar2024high,ji2024language, zhang2024speechtokenizer,kim2024clam,langman2024spectral}. These audio codes can then serve as representation for downstream audio tasks \cite{Wang2023NeuralCL, wu2024codec,kim2024clam}. However, although these models retain high-fidelity information about acoustic details due to the reconstruction objective, learning the appropriate acoustic invariances remains a challenge \cite{wu2024codec}. Further, high-fidelity signal reconstruction is unlikely to be a biologically plausible objective; instead, human speech perception mechanisms critically abstract away from the low-level acoustics and learn robust invariances over the audio signal
\cite{okada_hierarchical_2010,tang2017intonational}.
\textbf{A second popular approach is prediction-based modeling}, where models are trained to predict features derived from the raw waveform \cite{hsu2021hubert,chung_w2v-bert_2021,chen2022wavlm} or a time-frequency representation of audio \cite{michel2016blind,chung2019unsupervised, shain2020acquiring,chung2020vector}. These prediction-based speech models broadly fall into two categories: autoregressive models, which predict future frames \cite{michel2016blind,chung2019unsupervised, shain2020acquiring,chung2020vector}, and masked prediction models, which predict masked frames from surrounding frames \cite{chi2021audio,hsu2021hubert,chung_w2v-bert_2021} (analogous to the causal and bi-directional prediction approaches in language modeling). The learned representations are then applied to various downstream audio tasks, for instance language modeling \cite{hsu2021hubert,lakhotia2021generative,borsos2023audiolm,wu2023improving}. One of the most widely used predictive models is HuBERT \cite{hsu2021hubert}, which adapts the bi-directional BERT \cite{devlin_bert_2019} objective for speech representation learning using self-generated \textit{k}-means pseudolabels.
\textbf{A third common approach is contrastive learning}, in which frames from different audio samples are pushed together or pulled apart in the embedding space according to a specified objective \cite{oord2018representation,baevski_wav2vec_2020,Saeed2020ContrastiveLO,bhati2022unsupervised}. One popular model is wav2vec2 \cite{baevski_wav2vec_2020} which contrasts masked-out audio segments from distractors in combination with an auxiliary objective. Although the contrastive approach can yield powerful representations, it requires heuristics to define positive and negative samples, implicitly enforcing which aspects of the audio signal are retained---potentially building in incorrect assumptions. Moreover, contrastive objectives often rely on directly contrasting embeddings across hundreds or thousands of samples simultaneously, which arguably is not a biologically plausible operation. 

Although these three speech representation learning strategies are distinct, their objectives can be combined and augmented with additional heuristics. For instance, a state-of-the-art model, WavLM \cite{chen2022wavlm}, combines the HuBERT prediction objective \cite{hsu2021hubert} with a noisy input transformation. However, as with most ensemble models, these performance gains come at the cost of additional hand-crafted complexity.

\subsection{Our Approach: A Two-Stage Framework for Autoregressive Prediction on Biologically-Inspired Inputs} 
\label{intro:ours}
\begin{figure}
\centering
\includegraphics[width=1.0\linewidth]{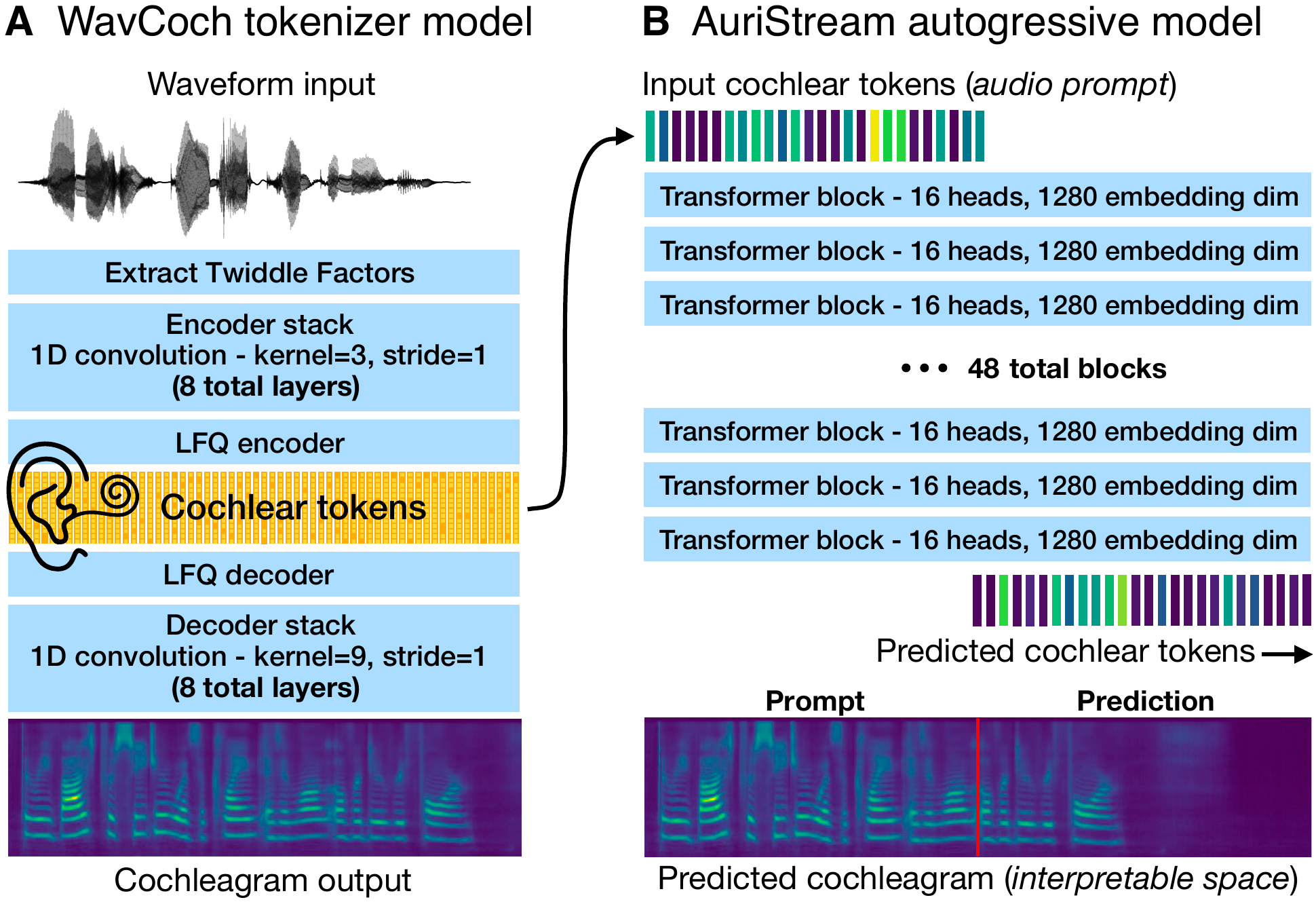}
\caption{\textbf{Schematic of speech representation framework}. Description of the steps of the framework can be found in the Introduction (\ref{intro:ours}) and Methods (\ref{methods:framework}).}
\label{fig:schematic}
\end{figure}

Unlike past approaches, our proposed framework does not rely on signal-reconstruction objectives (used by neural codec models), non-causal prediction objectives (used in bi-directional prediction models), or intra-batch contrasting of samples (used in many contrastive models). Instead, our framework takes inspiration from the human auditory processing hierarchy and operates in two stages: 

The first stage is \textbf{WavCoch}, a model that transforms the raw audio into a time-frequency representation based on the human cochlea (Figure \ref{fig:schematic}A). This approach bears some resemblance to neural audio codecs; however, instead of reconstructing the \textit{same} signal, we predict a \textit{different} audio representation--the time-frequency cochleagram--one known to be computed within the human auditory processing hierarchy \cite{wang1994self,cognitiveneuro,chi_multiresolution_2005,feather2023model}. We read out the representations from an intermediate bottleneck stage of WavCoch which effectively discretizes the audio representations. We refer to these intermediate representations as \textbf{cochlear tokens} (Figure \ref{fig:schematic}A). 

The cochlear tokens serve as input to the second stage, \textbf{AuriStream}, which is an autoregressive sequence model, trained to predict the upcoming cochlear tokens (Figure \ref{fig:schematic}B). Because the cochlear tokens were derived from a waveform-to-cochleagram transformation, these predicted tokens can naturally be decoded into the cochleagram, and then into audio, enabling inspection and interpretability. 

In sum, we formulate speech representation learning as a simple yet powerful autoregressive prediction task over biologically-realistic inputs---cochlear tokens. Our framework yields representations from which phonemes, word forms, and word meanings (lexical semantics) can be decoded at competitive levels (Section \ref{results:repr}), achieving state-of-the-art performance on lexical semantics. The learned representations also serve as a powerful backbone for various downstream SUPERB speech tasks \cite{yang2021superb} (Section \ref{results:superb}). Finally, unlike comparison models, AuriStream \textit{generates} continuations of audio that can be visualized in a cochleagram space, offering insights into the model's predictions (Section \ref{section:rollouts}).

\section{Methods}
\label{methods:framework}

\subsection{Input Tokens: WavCoch}
\label{methods:wavcoch}
We propose \textbf{WavCoch}, a model that efficiently tokenizes audio by transforming waveforms into cochleagrams, loosely mimicking the function of the human cochlea \cite{wang1994self,feather2023model}. The purpose of WavCoch is to extract discrete tokens from continuous audio signals to serve as the input to AuriStream.
WavCoch is a causal encoder-decoder network with 8 encoder layers (1D convolution with kernel 3) and 8 decoder layers (1D convolution with kernel 9) with a total of 11.1M parameters (see Figure \ref{app:architecture}A). It takes as input 5s clips of mono audio waveforms sampled at 16kHz and is trained to predict the cochleagram representation of this audio clip. The target cochleagram \cite{glasberg_derivation_1990,mcdermott_sound_2011,feather2023model} consists of 211 frequency bins and 988 temporal steps \cite{feather2023model} (for comparison to mel-spectrograms, see Appendix \ref{app:AuriStreamcochleagram-vs-mel-spectrogram}). To obtain discrete representations, we place a 13-bit LFQ \cite{MagViT2} bottleneck layer in the middle of the model, discretizing the embeddings into one of 8,192 units (corresponding to a 13-bit code, $2^{13}$) denoted as \textbf{cochlear tokens}.
We train WavCoch on the 960-hour LibriSpeech \cite{panayotov2015librispeech} dataset for 200k steps using the AdamW optimizer with a peak learning rate of 1e-4 with a 2,000-step warmup, and a cosine decay schedule. For further details and vocabulary size ablations, see Appendix \ref{app:architecture} and \ref{app:wavcoch-vocab-size-ablation}.

\subsection{Sequence Modeling: AuriStream}
\label{methods:seqmodel}
\textbf{AuriStream} is a GPT-style autoregressive Transformer \cite{radford_improving_2018} trained to predict the next cochlear token in a sequence (see Figure \ref{app:architecture}B). We train two versions: AuriStream-100M (100.7M parameters), with 12 layers, 12 attention heads and an embedding size of 784; and AuriStream-1B (970.1M parameters) with 48 layers, 16 attention heads, and an embedding size of 1,280. Both use SiLU activations \cite{hendrycks2016gelu}, RMSNorms \cite{zhang2019rmsnorm}, and a vocabulary of 8,192 cochlear tokens. The AuriStream model takes as input the cochlear token sequence produced by WavCoch and predicts the next token in the sequence using a context window of 4,096 tokens (approximately 20s of speech). We utilize a learned positional embedding and compute the cross-entropy loss between the predicted logits and the true next token in the sequence. We train both AuriStream models on the 60k hour LibriLight \cite{Kahn2019LibriLightAB} dataset for 500k steps using the AdamW optimizer with a peak learning rate of 3e-4 with a 2,000-step warmup, and a cosine decay schedule.

\subsection{Evaluation Metrics}
\subsubsection{Phoneme/Word Linear Probing}
\label{methods:probing}
To probe for phoneme and word identity representation, we use the TIMIT dataset \cite{garofolo1993timit} consisting of approximately five hours of audio recordings with ground truth phoneme- and word-boundaries. We use the train and complete test sets with exclusion of the ``SA'' sentences for train and test sets that are non-overlapping in sentences and speakers. For phoneme classification, we followed the standard protocol of collapsing the TIMIT phoneme labels from 60 to 39 classes \cite{lee1989speaker}. We embed the audio clip up to and including the target phoneme/word jointly, then extract just the embeddings of the time bins corresponding to that phoneme/word for probing.
We use the scikit-learn LogisticRegression multiclass classifier \cite{pedregosa_scikit-learn_2011}. The reported values are weighted accuracy scores as the classes are imbalanced. 

\subsubsection{Lexical Semantic Similarity (sSIMI)}
\label{methods:ssimi}
We use the ``sSIMI'' lexical semantics benchmark developed for the ZeroSpeech 2021 challenge \cite{nguyen2020zero}. The benchmark consists of pairs of words with ground truth human similarity judgments (on a 0 and 10 scale) collected from behavioral experiments. 
For instance, a pair of words such as ``water'' and ``river'' have a human similarity score of 9.8, while a pair like ``festival'' and ``whiskers'' have a score of 0.2. 
The benchmark contains two audio subsets: i) a natural subset with word pairs present in LibriSpeech \cite{panayotov2015librispeech}, and ii) a synthetic subset with all pairs. The sSIMI score is computed as the Spearman correlation between the cosine distance of model embeddings for word pairs and the true human similarity scores, multiplied by 100.

\subsubsection{Obtaining Model Embeddings}
\label{methods:embedding}
We obtain model embeddings for phoneme/word probing (Section \ref{methods:probing}) and lexical similarity (Section \ref{methods:ssimi}) by pooling the embeddings of all the tokens associated with the corresponding temporal section of the audio via ground-truth phoneme or word boundaries. For the pooling operation, we tested mean/max/min pooling across the temporal dimension. To select the best layer for decoding, we evaluate the phoneme/word probing performance on a subset of the TIMIT set (the top 10 phonemes/words in the TIMIT test set). For the sSIMI benchmark, we select the best layer on the independent ``dev'' set.


\subsubsection{Speech processing Universal PERformance Benchmark (SUPERB)}
\label{methods:superb}
We evaluate AuriStream on the SUPERB benchmark which contains 15 tasks, categorized into five aspects of speech: content, speaker, semantics, paralinguistics, and generation. We report values on a subset of six tasks spanning all five categories. We refer to the original paper for additional details on the benchmark \cite{yang2021superb}. Scores for the comparison models were obtained from the SUPERB leaderboard.

\section{Results}
\label{sec:results}

\subsection{AuriStream Embeddings Contain Information about Phoneme Identity, Word Identity, and Lexical Semantics}
\label{results:repr}
To first assess whether AuriStream representations contain information about phoneme and word identity, we trained linear classifiers on the phonemes and words from the TIMIT train set \cite{garofolo1993timit} and evaluated the classifiers on the test set with non-overlapping sentences and speakers. We compared AuriStream to five state-of-the-art speech representation models (see details in Appendix \ref{methods:comparisons}). As shown in Table \ref{tbl:probe}, for phoneme decoding, AuriStream-1B's performance was very close to state-of-the-art models HuBERT-xl and WavLM-large. The error patterns of AuriStream were sensible. For instance, the phoneme cluster ``er'' was often confused with ``r'', or ``ah'' with ``ih'' (see Appendix \ref{app:phoneme}). 
For word decoding, AuriStream-1B surpassed wav2vec-large, however, AuriStream fell short of HuBERT and WavLM. We hypothesize that the subpar word decoding performance of AuriStream relative to these models is due to the fact that HuBERT and its derivative models (WavLM) were exposed to global clustering operations aimed at discovering word-like units. In contrast, AuriStream did not undergo any such global operations.
Finally, we emphasize that decoding performance for both phonemes and words scales well with AuriStream size.

\begin{table}[t]
  \caption{\textbf{Linear probing performance for phonemes or words on the TIMIT dataset}. Reported values are weighted accuracy scores of the best layer (see Section~\ref{methods:embedding}) on the TIMIT test set with non-overlapping sentences uttered by non-overlapping speakers relative to the train set.}
  \centering
  \setlength{\tabcolsep}{3.5pt}
  \begin{tabular}{lcccc}
    \toprule
    Dataset          & Params & Hours & Phoneme & Word \\
    \midrule
    HuBERT-base      & 97M    & 1K    & 0.83    & 0.75 \\
    HuBERT-xl        & 1000M  & 60K   & \textbf{0.91} & \textbf{0.85} \\
    wav2vec2-large   & 317M   & 60K   & 0.76    & 0.43 \\
    WavLM-base       & 97M    & 1K   & 0.85 & 0.76 \\
    WavLM-large      & 317M   & 94K   & 0.90    & 0.84 \\
    AuriStream-100M  & 101M   & 60K   & 0.82    & 0.45 \\
    AuriStream-1B    & 970M  & 60K   & 0.88    & 0.65 \\
    \bottomrule
  \end{tabular}
  \label{tbl:probe}
\end{table}

Second, we evaluate whether AuriStream learns representations of word meanings (lexical semantics). This benchmark (sSIMI) measures the correlation between embeddings of audio corresponding to pairs of words (e.g., ``water'' and ``river'') and human similarity judgments. Prior studies have described speech models' performance on this task as ``modest'' \cite{nguyen2020zero}. As shown in Table \ref{tbl:ssimi}, both AuriStream-100M and AuriStream-1B outperform the other models on the natural and synthetic data subsets, and AuriStream-1B outperformed all other models on the synthetic set. Performance also improves with model scale. These findings demonstrate that a simple autoregressive prediction objective can lead to state-of-the-art representations for lexical semantics. 

\begin{table}[t]
  \caption{\textbf{Semantic similarity scores on the ZeroSpeech 2021 Lexical Semantic Benchmark}. Reported values are Spearman correlations (multiplied by 100 per \cite{nguyen2020zero}) of the best layer (see Section~\ref{methods:embedding}) between the embeddings for pairs of words and human similarity judgments. Scores are obtained on the test sets of two subsets: LibriSpeech Audio and Synthetic Audio.}
  \centering
  \setlength{\tabcolsep}{3.5pt}
  \begin{tabular}{lcccc}
    \toprule
    Dataset         & Params  & Hours & LibriSpeech $\uparrow$ & Synthetic $\uparrow$ \\
    \midrule
    HuBERT-base     & 97M     & 1K    & 6.10  & 7.48  \\
    HuBERT-xl       & 1000M   & 60K   & 7.81  & 10.37 \\
    wav2vec2-large  & 317M    & 60K   & 6.41  & 7.19  \\
    WavLM-base      & 97M     & 1K    & 8.29  & 9.41 \\
    WavLM-large     & 317M    & 94K   & 10.50 & 10.37 \\
    AuriStream-100M & 101M    & 60K   & 10.63 & 10.12 \\
    AuriStream-1B   & 970M    & 60K   & \textbf{12.52} & \textbf{10.64} \\
    \bottomrule
  \end{tabular}
  \label{tbl:ssimi}
\end{table}

\subsection{AuriStream Serves as a Strong Backbone for Downstream Audio Tasks}
\label{results:superb}

\begin{table}[t]
  \centering
  \caption{\small \textbf{Model performance on SUPERB tasks}. Reported values are obtained by training a downstream task decoder on top of a frozen model backbone \cite{yang2021superb}. ASR = automatic speech recognition, IC = intent classification, KS = keyword spotting, SID = speaker identification, ER = emotion recognition, SS = speaker separation.}
  \setlength{\tabcolsep}{3.5pt}
  \begin{tabular}{lccccccc}
    \toprule
    Setting         & ASR $\downarrow$ & IC $\uparrow$ & KS $\uparrow$ & SID$\uparrow$ & ER $\uparrow$ & SS $\uparrow$ \\
    \midrule
    HuBERT-base     & 6.42  & 98.34 & 96.30 & 81.42 & 64.92 & 9.36 \\
    HuBERT-large    & 3.62  & 98.76 & 95.29 & 90.33 & 67.62 & 10.45 \\
    wav2vec2-large  & 3.75  & 95.28 & 96.66 & 86.14 & 65.64 & 10.02 \\
    WavLM-base      & 6.21  & 98.42 & 96.79 & 84.51 & 65.94 & 10.37 \\
    WavLM-large     & \textbf{3.44}  & \textbf{99.31} & \textbf{97.86} & \textbf{95.49}  & \textbf{70.62} & \textbf{11.19} \\
    vq-wav2vec      & 17.71 & 85.68 & 93.38 & 38.80 & 58.24 & 8.16  \\
    APC             & 21.28 & 74.69 & 91.01 & 60.42 & 59.33 & 8.92  \\
    AuriStream-100M & 7.80  & 92.00 & 93.96 & 79.10 & 59.32 & 9.05  \\
    AuriStream-1B   & 4.20  & 98.01 & 95.25 & 81.14 & 67.47 & 10.07 \\
    \bottomrule
  \end{tabular}
  \label{tbl:superb}
\end{table}

Having established that AuriStream representations encode meaningful phoneme, word, and lexical semantics information, we investigated whether the frozen representations of AuriStream would serve as powerful features for training decoders across various audio tasks. To do so, we leveraged six tasks from the SUPERB benchmark, spanning all five major task categories defined in the benchmark \cite{yang2021superb}.  
As shown in Table \ref{tbl:superb}, AuriStream-1B outperformed APC and vq-wav2vec---two models most similar to AuriStream---while performing competitively against state-of-the-art models on most tasks. In particular, AuriStream-1B showed strong performance on automatic speech recognition (ASR), intent classification (IC), and speech separation (SS). In contrast, AuriStream-1B had subpar performance on keyword spotting (KS) compared to other similarly sized models. Although WavLM-large---a model which contains many hand-designed heuristics such as noise addition during training and \textit{k}-means clustering---dominates in all categories, AuriStream comes close to matching its performance on several tasks, demonstrating that it learns versatile representations for diverse downstream audio tasks. Importantly, AuriStream's favorable scaling behavior indicates strong potential for further improvements with more parameters and training data.

\subsection{AuriStream Learns Short- and Long-Range Speech Statistics}
\label{section:rollouts}
\begin{figure}[t]
\centering
\includegraphics[width=\linewidth, trim=10 10 10 10, clip]{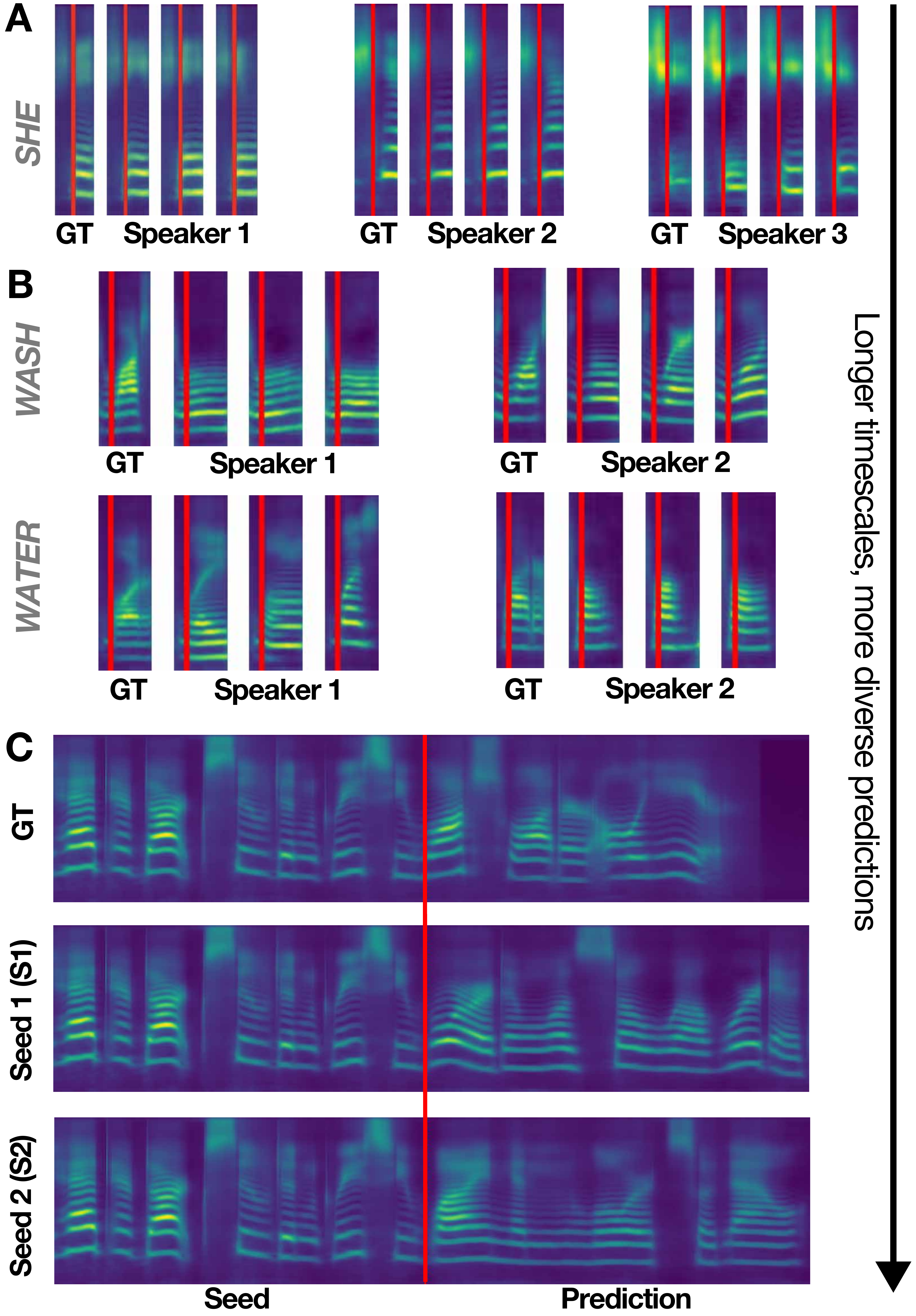}
\caption{\textbf{Cochleagram predictions by AuriStream-1B.} \textbf{A.} AuriStream-1B is prompted with the first phoneme of the word ``she'' (left of red vertical line) and predicts the word completion (right of red line) across three seeds. The ground-truth (GT) cochleagram is shown in the first column. \textbf{B.} AuriStream is prompted with the first phoneme of the words ``wash'' and ``water''. \textbf{C.} AuriStream is prompted with the first 2.5 seconds of an audio clip (red line) from the TIMIT test set and predicts the remaining part of the clip across two different seeds. }
\vspace{-8pt}
\label{fig:rollout+plot}
\end{figure}

In this final section, we leverage the fact that AuriStream was trained to perform predictions in a space that can be visualized and interpreted (the time-frequency cochleagram image) to ask whether it learns speech statistics without ground-truth phoneme, word, or task labels.
We hypothesize that learned speech statistics should manifest in two distinct modes: \textit{At short timescales}, when provided with sufficient context (such as the first part of a common word), the model should complete the cochleagram in a way that aligns with the remainder of the word. In contrast, \textit{at longer timescales}, the model's predictions should diverge, reflecting the variability of plausible words that could follow any given phoneme or word. 
To test this hypothesis, we provided AuriStream with variable-length sequences of ground-truth audio clips from the TIMIT test set (out-of-distribution for AuriStream) and qualitatively analyzed the resulting model completions (Figure \ref{fig:rollout+plot}).

To first test whether AuriStream learns speech structure at short timescales, we prompted the model with the first phoneme of a common word (e.g., ``she'', starting with the phoneme ``sh''), and evaluate predictions from this first phoneme across different speakers in the TIMIT test set. As shown in Figure \ref{fig:rollout+plot}A, the model learns to consistently complete the phoneme with an ``iy'' phoneme, resulting in the word ``she''. Conversely, when a phoneme has several likely continuations, the model learns to complete the phoneme with different words. For instance, when prompted with the initial phoneme cluster (``wa'') of the words ``water'' and ``wash'' from two different speakers (Figure \ref{fig:rollout+plot}B), AuriStream sometimes predicts the remainder of the true word and other times generates a different completion consistent with the initial phoneme cluster. In one example, AuriStream's prediction for Speaker 2's utterance ``wash'' appears more similar to Speaker 1's ground-truth utterance of the word ``water'' than to its own ground-truth word (``wash'') (Figure \ref{fig:rollout+plot}B), indicating that AuriStream learns to complete phoneme prompts with different plausible word continuations. These visualizations suggest that AuriStream learns the statistical regularities of how phonemes combine to form words, demonstrating knowledge of speech structure at short timescales. 

Second, to evaluate the diversity of longer-range predictions, we prompted AuriStream with the first 2.5 seconds of TIMIT audio clips (Figure \ref{fig:rollout+plot}C). AuriStream predicts several seconds of plausible continuations as inspected visually, and audibly, since cochleagram representations can be inverted into audio. These continuations often sound very plausible given the topic of the prompt. We observe that the continuations degrade over time, however, we emphasize that the purpose of AuriStream is not to be a \textit{language} model, but a \textit{speech} representation model---the fact that it can perform rudimentary language modeling is a serendipitous side effect of the training objective, which points to the fact that learning patterns in speech, and producing language may be operationalized under a unified objective, albeit perhaps requiring additional mechanisms for enforcing longer-range coherence.
Additional audio samples available at: \href{https://tukoresearch.github.io/auristream-speech/}{{https://tukoresearch.github.io/auristream-speech/}} (also see details in Appendix \ref{app:interpretability}).

\section{Conclusion}
\label{sec:disc}
We introduced AuriStream, a self-supervised speech representation model that achieves competitive phoneme and word decoding, state-of-the-art lexical semantic representations, and serves as a strong representational backbone for various audio tasks.
A key strength of our framework is the use of cochlear tokens: a biologically inspired and highly efficient token representation (around 200 tokens per second of audio) that fits within the context window of a standard Transformer, effectively leveraging the power of autoregressive modeling. While WavCoch is conceptually similar to neural codec approaches \cite{Zeghidour2021SoundStreamAE,yang2023hifi,ji2024language, zhang2024speechtokenizer,kim2024clam}, its novelty lies in learning to transform one representation into \textit{another} representation through a discrete quantization bottleneck (instead of auto-encoding, as done in related approaches)---an approach we denote as ``Transformation Imitation''.
Finally, unlike prior speech representation models such as HuBERT and wav2vec2 \cite{hsu2021hubert,baevski_wav2vec_2020}, AuriStream can also \textit{generate} audio, enabling both embedding extraction and audio generation. In addition, AuriStream enables the visualization and interpretation of audio predictions through the cochleagram space, a capability that many audio models lack, making AuriStream less of a ``black box''. 

Limitations of our work exist. One limitation is that AuriStream is trained on English speech, restricting analyses to tasks and materials in English \cite{blasi2022over,feng2024towards}. Another limitation is that AuriStream is trained exclusively on read speech from LibriLight, limiting ecological validity. Extending training to more naturalistic and developmentally plausible data \cite{sullivan2021saycam,warstadt_call_2023} is a key future direction.
More broadly, although AuriStream is not a fully biologically realistic model, it constitutes a critical step in the right direction---and we hope that it will serve as a valuable model for the emerging field of ``NeuroAI'' which aims to understand biological and artificial intelligence by linking the representations and computations in artificial models to neural activity in the brain \cite{kell_task-optimized_2018,millet_toward_2022,li_dissecting_2022,tuckute2023many,oota2023speech,tuckute2024language,moussa2025brain}. 

\section{Acknowledgements}
G.T. acknowledges support from The K. Lisa Yang ICoN Center and McGovern Institute for Brain Research. E.F. acknowledges support from McGovern Institute for Brain Research, the Department of Brain and Cognitive Sciences, MIT's Quest for Intelligence, and the Simons Foundation. K.K. and D.L.K.Y. acknowledge support from the Simons Foundation (543061), National Science Foundation (CAREER grant 1844724), Office of Naval Research (MURI S5847 and 1141386 - 493027).
We thank the Stanford HAI, Stanford Data Sciences and the Marlowe team, and the Google TPU Research Cloud team for computing support.

\bibliographystyle{IEEEtran}
\bibliography{main}

\clearpage

\newpage

\section{Appendix}

\setcounter{figure}{0}
\setcounter{table}{0}
\renewcommand{\thefigure}{\Roman{figure}}
\renewcommand{\thetable}{\Roman{table}}

\subsection{WavCoch Architecture Details}
\label{app:architecture}

As shown in Figure \ref{fig:schematic}A, the raw waveform (shape: 1 $\times$ 80,000 for 5s of mono audio sampled at 16kHz) is first transformed into the time–frequency domain via a fixed-kernel discrete Fourier transform implemented as a bank of 1D convolutional filters (window size 1,001 samples, hop length 80 samples). The filter weights---the complex sinusoidal basis functions (or Twiddle Factors \cite{Cooley1965AnAF}) of the discrete Fourier transform---slide over the signal to produce a spectral representation with one feature vector every 5ms.
Second, each 5ms temporal step of this frequency representation is passed through an 8-layer encoder stack (each layer is a 1D convolution with 512 channels, kernel size 3, stride 1, ReLU nonlinearities), yielding a sequence of 512-dimensional embeddings.
Third, these embeddings are then passed through a 13-dimensional LFQ bottleneck \cite{MagViT2}, which effectively binarizes the representation. We read out the activations of this bottleneck as a 13-bit binary code which can be interpreted as one of $2^{13} = 8,192$ discrete tokens. We determined that 13-bits is the optimal vocabulary size by ablating vocabulary sizes and evaluating out-of-distribution performance on cochleagram reconstruction error and phoneme cluster purity; 12-bit and 14-bit codes yielded inferior performance (see full ablation details in Appendix \ref{app:wavcoch-vocab-size-ablation}).
Fourth, the output of the LFQ bottleneck is passed through a decoder stack (each layer is a 1D convolution with 211 channels, kernel size 9, stride 1, ReLU nonlinearities). This decoder output corresponds to the frequencies in the cochleagram representation \cite{feather2023model}, which the model is supervised to match via L2 error. An auxiliary entropy penalty with a weight of 0.001 is applied at the LFQ bottleneck to encourage diversity, in line with \cite{MagViT2}.
Thus, for every 5 seconds of audio, WavCoch extracts a sequence of 988 integers in the range [0, 8192) through the LFQ bottleneck, denoted as cochlear tokens, to feed into AuriStream (illustrated in Figure \ref{fig:schematic}B).

\subsection{WavCoch Vocabulary Size Ablations}
\label{app:wavcoch-vocab-size-ablation}
\begin{figure}[!htb]
\centering
\includegraphics[width=\linewidth]{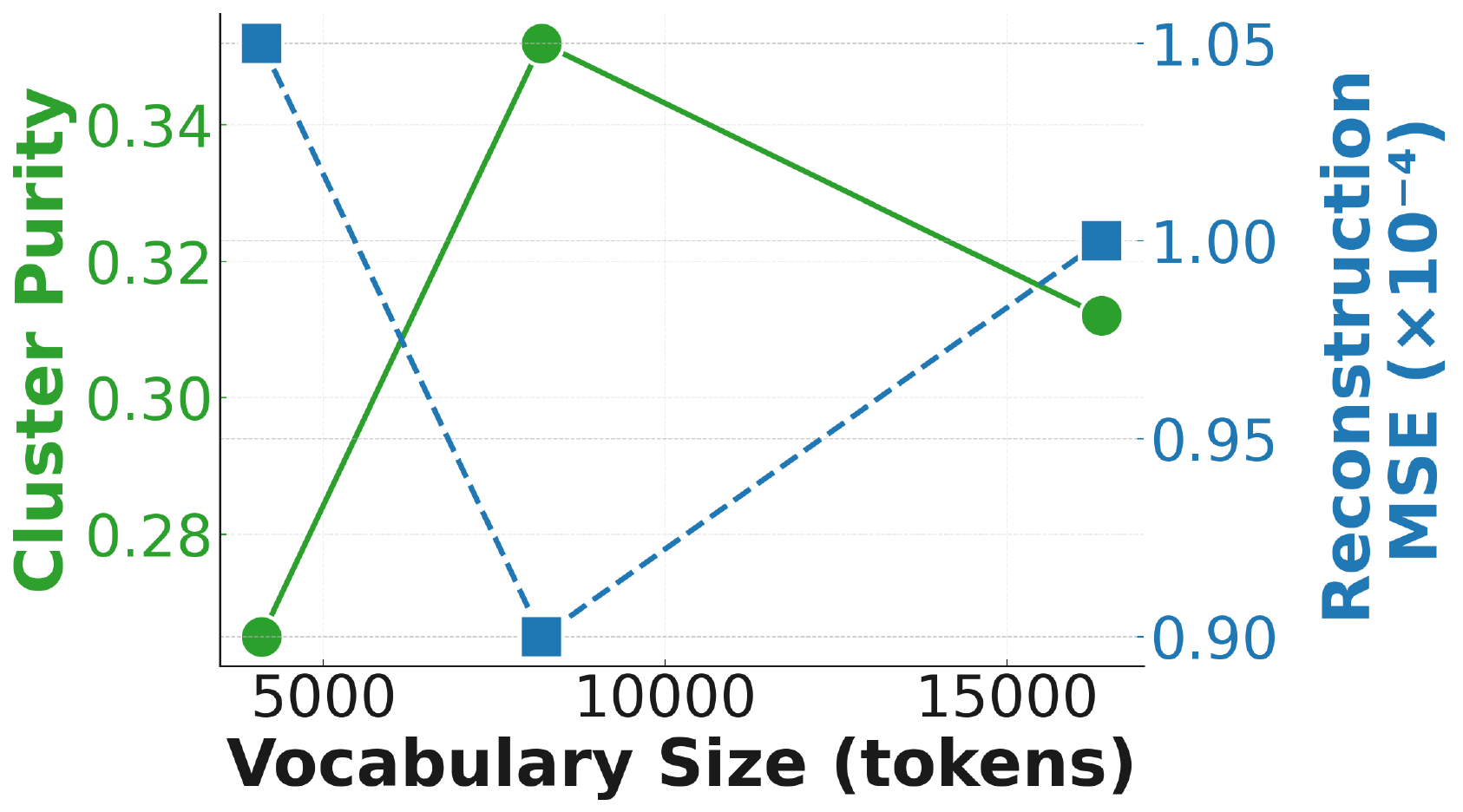}
\caption{\textbf{Evaluation of WavCoch trained with different vocabulary sizes.} We plot the L2 cochleagram reconstruction error (blue) and the phoneme cluster purity (green) on the out-of-distribution TIMIT test set.}
\label{appendix:wavcoch-vocab-size-ablation}
\end{figure}

We performed ablations to identify the optimal vocabulary size of the WavCoch model. We trained variants of WavCoch using a vocabulary size of 4,096, 8,192, and 16,384 (12-, 13- and 14-bit codes, respectively) on the LibriSpeech960 dataset \cite{panayotov_librispeech_2015}. For each of these models, we evaluated the cochleagram reconstruction L2 error and phoneme cluster purity on an out-of-distribution test set (TIMIT test set \cite{garofolo1993timit}). Phoneme cluster purity was defined as \textit{purity = (count of most associated phoneme for token i) / (total counts for token i)} providing an intuitive metric for how consistently a given token aligns with a specific phoneme. 
Figure \ref{appendix:wavcoch-vocab-size-ablation} shows that a vocabulary size of 8,192 (13-bit code) yields both the lowest reconstruction error and the highest phoneme cluster purity. 

\subsection{WavCoch Target Representations: Cochleagram vs. Mel Spectrogram}
\label{app:AuriStreamcochleagram-vs-mel-spectrogram}
To evaluate the impact of using the biologically-inspired cochleagram representation \cite{wang1994self,feather2023model} as the WavCoch prediction target as opposed to the more standard deep learning practice of using a mel-spectrogram, we trained a version of WavCoch using mel-spectrograms (80 mel bins and 5ms temporal bins) as prediction targets. Both cochleagram- and mel-based WavCoch models were trained on the publicly available LibriSpeech960 dataset \cite{panayotov_librispeech_2015}, consisting of 960 hours of speech recordings. 
Since the L2 reconstruction error is not directly comparable between a cochleagram and a mel-spectrogram, we investigated two proxy measures of representational quality: i) The number of unique codes utilized in the quantized representation (``codebook usage''), and ii) Phoneme cluster purity (defined as \textit{purity = (count of most associated phoneme for token i) / (total counts for token i)}) Both metrics were computed on the out-of-distribution TIMIT test set \cite{garofolo1993timit} and are reported in Table \ref{tbl:coch-vs-mel}.

\setlength{\tabcolsep}{3.5pt}
\begin{table}[t]
\centering
\caption{\small \textbf{Evaluation of WavCoch trained with different prediction targets}. Codebook usage and phoneme cluster purity evaluated on the out-of-distribution TIMIT test set.}
\vspace{5pt}
\setlength{\tabcolsep}{3.5pt}
    \begin{tabular}{lccc}
    \toprule
    Target & Codebook Usage $\uparrow$  & Cluster Purity $\uparrow$ \\
    \midrule
    Cochleagram & \textbf{8,172} & \textbf{0.3517}  \\
    Mel-Spectrogram & 8,151 & 0.3473 \\
    \bottomrule
    \end{tabular}
\label{tbl:coch-vs-mel}
\end{table}

First, in terms of codebook usage, we found that the WavCoch model trained with the cochleagram target utilized slightly more codes than the model trained with the mel-spectrogram target to represent out-of-distribution speech data (TIMIT test \cite{garofolo1993timit}). Second, the cochleagram-based WavCoch model achieved a slightly higher average phoneme cluster purity on the TIMIT test set than the mel-spectrogram model. While these differences are relatively small, they suggest that the cochleagram representation performs at least as well as, if not slightly better than the mel-spectrogram in this setting. 

Beyond the quantitative analyses reported in Table \ref{tbl:coch-vs-mel}, we prefer the cochleagram over the mel-spectrogram representation for conceptual reasons: The ultimate goal of our framework is to move towards more biologically plausible speech models, and the cochleagram is more aligned with this goal.

\subsection{Comparison Models}
\label{methods:comparisons}
AuriStream is compared to five state-of-the-art speech representation models using the HuggingFace Transformers package: HuBERT-base (identifier: \textit{facebook/hubert-base-ls960}), HuBERT-xl (identifier: \textit{facebook/hubert-xlarge-ll60k}), wav2vec2-large (identifier: \textit{facebook/wav2vec2-large}), WavLM-base (identifier: \textit{microsoft/wavlm-base}), and WavLM-large (identifier: \textit{microsoft/wavlm-large}). For the SUPERB benchmark, we additionally compare against two smaller models which share some similarity to AuriStream, specifically, APC and vq-wav2vec.

\subsection{Confusion Matrix for Phoneme Decoding}
\label{app:phoneme}
\begin{figure}[!htb]
\centering
\includegraphics[width=\linewidth]{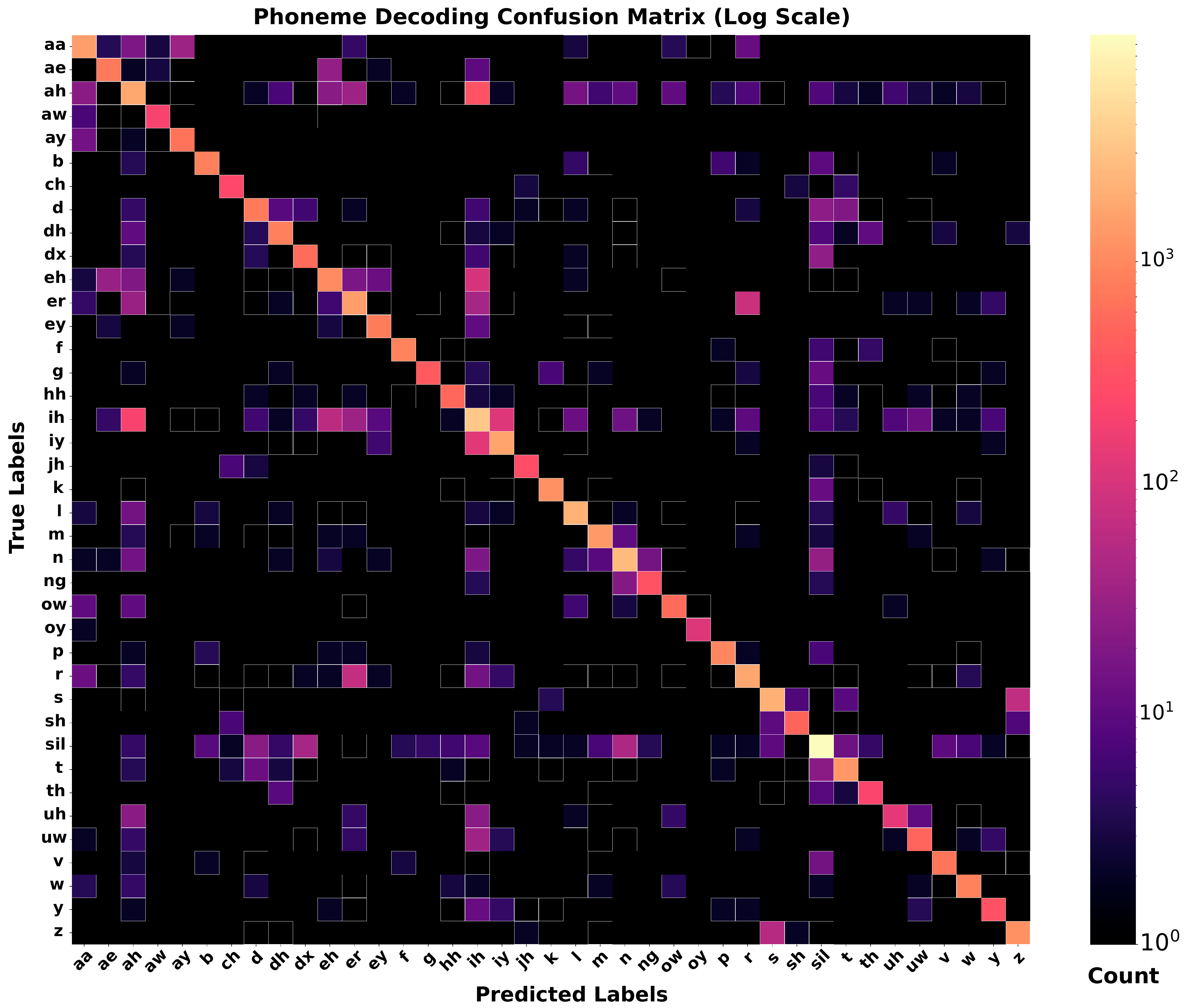}
\caption{\textbf{Confusion matrix for phoneme decoding.} The plot shows which phonemes were confused with each other from the AuriStream-1B model on the TIMIT test set. The plot is shown on a log colorscale to better highlight the mismatches between true and predicted labels.}
\label{appendix:phoneme}
\end{figure}

Figure \ref{appendix:phoneme} shows the phoneme confusion matrix for AuriStream-1B in the linear decoding task (see Section \ref{results:repr}). The error patterns were sensible: for instance, ``er'' was often confused with ``r', or ``ah'' with ``ih''.

\subsection{Sonifying AuriStream Predictions through Cochleagram Inversion}
\label{app:interpretability}
We investigate AuriStream's predictions by inverting the cochleagrams into audible waveforms. To this end, we developed a simple per-sample optimization procedure that constructs a waveform that matches the cochleagram prediction. 
Specifically, we optimize a tensor of shape (1 $\times$ 80,000)---initialized with random numbers from a normal distribution with mean 0 and variance 1---representing the waveform input to make its cochleagram representation match the cochleagram predicted by WavCoch (via L2 error). We backpropagate through the cochleagram transformation and use the Adam optimizer with a learning rate of 1e-2.
Note that this optimization procedure is not a learned vocoder model, but a simple procedure which converts the output of WavCoch, the cochleagrams, into audible sound (conceptually similar to Griffin-Lim algorithm).

Several audible samples of speech generations from AuriStream-1B are available at the following link: \href{https://tukoresearch.github.io/auristream-speech/}{https://tukoresearch.github.io/auristream-speech/}. Please access the page using Google Chrome as we have seen some cases in which Safari and Firefox are not properly loading these videos.

We observed that on short timescales, the model produces reasonable completions, but the longer the completion, the more the predictions drift away from being plausible. We want to emphasize that the purpose of AuriStream is not to be a language model, but a speech representation model---the fact that it can perform rudimentary language modeling is a serendipitous side effect of the training objective, which points to the fact that learning patterns in speech, and producing language may be operationalized under a unified objective. These findings serve as great motivating factors for follow-up work, which will attempt to stabilize speech generations with longer-term coherence, building on the foundation laid out in this paper.

\end{document}